\documentclass[letterpaper]{article} 
\usepackage{aaai25}  
\nocopyright 
\usepackage{times}  
\usepackage{helvet}  
\usepackage{courier}  
\usepackage[hyphens]{url}  
\usepackage{graphicx} 
\usepackage{natbib}  
\usepackage{caption} 
\usepackage{algorithm}
\usepackage{algorithmic}

\usepackage{newfloat}
\usepackage{listings}
\DeclareCaptionStyle{ruled}{labelfont=normalfont,labelsep=colon,strut=off} 
\floatstyle{ruled}
\newfloat{listing}{tb}{lst}{}
\floatname{listing}{Listing}
\usepackage{amsmath,amssymb,amsfonts}
\usepackage{algorithmic}
\usepackage{graphicx}
\usepackage{textcomp}
\usepackage{xcolor}
\def\BibTeX{{\rm B\kern-.05em{\sc i\kern-.025em b}\kern-.08em
    T\kern-.1667em\lower.7ex\hbox{E}\kern-.125emX}}

\usepackage[utf8]{inputenc} 
\usepackage{url}            
\usepackage{booktabs}       
\usepackage{makecell}       
\usepackage{amsfonts}       
\usepackage{nicefrac}       
\usepackage{microtype}      
\usepackage{xcolor}         

\usepackage{mathtools}
\usepackage{amsthm}
\usepackage{thmtools}
\usepackage{thm-restate}

\usepackage[capitalize,noabbrev,nameinlink]{cleveref}

\usepackage[]{caption}
\usepackage{subcaption}
\usepackage{graphicx,xcolor,graphics,url}
\usepackage{verbatim,ifthen,alltt,array}
\usepackage{latexsym,dsfont}
\usepackage{booktabs,multirow}
\usepackage{enumitem}
\usepackage{xparse}

\usepackage{xspace}

\usepackage{xfrac}
\usepackage[framemethod=TikZ]{mdframed}
\usepackage{nicefrac}       

\usepackage{stmaryrd}
\usepackage{textcomp}
\usepackage{siunitx}
\usepackage{xpatch}

\usepackage{pgf}
\usepackage{tikz}
\usepackage{forest}

\usetikzlibrary{arrows,decorations.pathmorphing,decorations.pathreplacing,decorations.footprints,fadings,calc,trees,mindmap,shadows,decorations.text,patterns,positioning,shapes,matrix,fit}
\usetikzlibrary{arrows,shadows,backgrounds}
\usetikzlibrary{arrows.meta}
\usetikzlibrary{positioning,fit}
\usetikzlibrary{automata}
\usetikzlibrary{matrix}
\usetikzlibrary{shapes.symbols,shapes.misc,shapes.arrows}
\usetikzlibrary{matrix,chains,scopes,decorations.pathmorphing}
\usetikzlibrary{bayesnet}
\usetikzlibrary{tikzmark}

\newtheoremstyle{nthmstyle}
{3pt}
{3pt}
{}
{}
{\bfseries}
{.}
{.5em}
{}

\theoremstyle{nthmstyle}

\newtheorem{prop}{Proposition}
\newtheorem{example}{Example}
\newtheorem{runexmpl}{Running example}

\DeclareMathOperator*{\im}{\msf{Im}}
\DeclareMathOperator*{\sv}{\msf{Sc}}

\newcommand{\mbb}[1]{\ensuremath\mathbb{#1}}
\newcommand{\msf}[1]{\ensuremath\mathsf{#1}}
\newcommand{\tn}[1]{\textnormal{#1}}
\newcommand{\cf}{\ensuremath\upsilon} 
\newcommand{\imn}[1]{\msf{Im}_{#1}}
\newcommand{\diff}[1]{\Delta_{#1}}
\newcommand{\coef}{\varsigma}
\newcommand{\acron}{UMI\xspace}
\newcommand{\acronpl}{UMIs\xspace}
\newcommand{\pred}{\msf{P}}

\newcommand{\msp}{\msf{MSP}}
\newcommand{\mbp}{\msf{MBP}}
\newcommand{\dset}{\msf{DSet}}
\newcommand{\winc}{\msf{WinC}}
\newcommand{\send}{\msf{STerm}}
\newcommand{\critic}{\msf{Crit}}

\begin{document}

\title{The Sets of Power}
\author {
    Joao Marques-Silva\textsuperscript{\rm 1},
    Carlos Menc\'ia\textsuperscript{\rm 2},
    Ra\'ul Menc\'ia\textsuperscript{\rm 2}
}
\affiliations {
    \textsuperscript{\rm 1}ICREA, University of Lleida, Spain\\
    \textsuperscript{\rm 2}University of Oviedo, Spain\\
    jpms@icrea.cat,  \{menciacarlos, menciaraul\}@uniovi.es
}

%

\maketitle

\begin{abstract}
  Measures of voting power have been the subject of extensive research
  since the mid 1940s. More recently, similar measures of relative
  importance have been studied in other domains that include
  inconsistent knowledge bases, intensity of attacks in
  argumentation, different problems in the analysis of
  database management, and explainability.
  This paper demonstrates that all these examples are instantiations
  of computing measures of importance for a rather more general
  problem domain.
  The paper then shows that the best-known measures of importance can
  be computed for any reference set whenever one is given a
  monotonically increasing predicate that partitions the subsets of
  that reference set. 
  As a consequence, the paper also proves that measures of importance
  can be devised in several domains, for some of which such measures
  have not yet been studied nor proposed.
  Furthermore, the paper highlights several research directions
  related with computing measures of importance.
\end{abstract}

\section{Introduction} \label{sec:intro}

The measure of voting power in assemblies of voters has attracted the 
interest of researchers since at least the work of L.~Penrose in
the 1940s~\cite{penrose-jrss46}, with important contributions in the
following decades~\cite{shapley-apsr54,banzhaf-rlr65}.
More recently, measures of importance have been studied in
other
domains, that include
inconsistent knowledge
bases~\cite{hunter-kr06,hunter-aij10,jabbour-ijcai23},
intensity of attacks in argumentation~\cite{amgoud-ijcai17},
set covering~\cite{gusev-omega20,gusev-ejor23},
database management%
~\cite{kimelfeld-sigmod23},
but also
explainable artificial intelligence
(XAI)~\cite{lundberg-nips17,izza-aaai24,lhams-corr24,lhms-corr24}.
Among these, the recent uses of measures of importance in XAI have
drawn significant interest, with important limitations being
uncovered in recent work~\cite{msh-cacm24,hms-ijar24}.

However, despite the growing number of domains where measures of
relative importance have been studied, in each case a dedicated
formulation has been proposed.
In turn, this does not reveal possible connections between uses of
measures of importance in different domains, nor does it suggest how
the same measures can be applied to other domains.
More importantly, as this paper underlines, measures of importance can
readily be envisioned in different practical applications, not being
apparent how such measures might be devised.
For example, in model-based diagnosis~\cite{reiter-aij87}, one may be 
interested in ranking the components of a system in terms of their
relevancy for some observed faulty behavior.
Similarly, in the case of inconsistent systems of linear
inequalities~\cite{vanloon-ejor81}, one may want to assign relative
importance (for inconsistency) to the inequalities.
In a more general setting, the computation and enumeration of minimal
sets (of a reference set $N$) over a monotone predicate (MSMP) has
made significant 
progress in recent
years~\cite{msjb-cav13,msjm-aij17,berryhill-sat18,bendik-lmcs22}. 
However,
no solution has been proposed to assign relative importance to the
elements of $N$.
As a result, as shown later in the paper, many more use cases of
measures of relative importance can be envisioned.

In contrast to earlier works, this paper proposes a different take on
devising measures of relative importance.
Concretely, the paper shows that the best-known measures of importance
can be computed for any reference set whenever one is given a
monotonically increasing predicate that partitions the subsets of 
that reference set. 
As a consequence, the paper also proves that measures of importance
can be devised in several domains, for some of which such measures
have not yet been studied nor proposed.
Although the observations made in the paper could be perceived as
plain by some, it is also the case that such observations are not  
readily apparent to many practitioners, as the rediscovery of the same
ideas in different settings demonstrates.  
Moreover, the paper also summarizes additional application domains,
for which the use of relative measures of importance is outlined.
The paper also glances through the exact computation of measures of
importance, as well as their approximation in practice. 
Finally, the paper highlights 
research directions related with computing measures of importance in
novel application domains.

\section{Preliminaries} \label{sec:prelim}

The notation used in the paper is adapted from the one used in several
earlier works~\cite{msjb-cav13,slaney-ecai14,msjm-aij17}.

\paragraph{Sets, predicates \& monotonicity.}
The following sets are assumed: (i) $N=\{1,2,\ldots,m\}$, the
set of elements that we consider; and (ii) $\mbb{B}=\{0,1\}$, denoting
the outcomes of predicates.
A \emph{predicate} $\pred$ is a
mapping from subsets of $N$ to $\mbb{B}$, $\pred:2^{N}\to\mbb{B}$. (As
standard in different domains, we will equate 0 with $\bot$ and 1 with
$\top$, with $0 < 1$.)
A predicate partitions 
the power set 
of a set into two sets, the subsets of elements for 
which it takes value 0 (if the predicate does not hold), and 
those for which it takes value 1 (if it holds). 

Let ${\bowtie}\in\{\le,\ge\}$. 
We say that a
predicate is \emph{monotone} if whenever $X,Y\subseteq{N},Y\subseteq{X}$,
then it is the case that $\pred(X)\bowtie\pred(Y)$.
A monotone predicate is increasing when ${\bowtie}={\ge}$, and
decreasing when ${\bowtie}={\le}$. Predicates are assumed \emph{not}
to be constant; hence, for a monotonically increasing predicate we
must have $\pred(\emptyset)=0$ and $\pred(N)=1$, and for a monotonically
decreasing predicate we must have $\pred(\emptyset)=1$ and
$\pred(N)=0$.
Throughout the paper, predicates are assumed to be monotonically
increasing.

Finally, given a set $N$ and a predicate $\pred$ defined on $N$, a
subset $S\subseteq{N}$ is minimal (with respect to $\pred$) if
$\pred(S)$ holds, and $\pred(T)$ does not hold for any proper subset
$T$ of $S$.

\paragraph{Graphs \& minimal hitting sets (MHSs).}
%
%
Let $N=\{1,\ldots,m\}$ denote a set of vertices. A graph $G$ is a
tuple $G=(N,E)$, where $E$ denotes a set of edges, consisting of a subset
of $\{\{i,j\}\,|\,i,j\in{N},i\not=j\}$. 
Notice we consider simple undirected graphs. 

Given a set of sets $S=\{S_1,\ldots,S_k\}$, a hitting set $H$ is a
set whose intersection with any of the sets in $S$ is not the empty set.
A minimal hitting set (MHS) is a hitting set such that none of its proper
subsets is a hitting set.

\paragraph{Minimal sets over a monotone predicate (MSMP).}
MSMP has been defined as the problem of computing a \emph{minimal}
subset $X$ of $N$, given a monotone predicate
$\pred:2^{N}\to\mbb{B}$, for which $\pred(X)$ holds. (In this context, $\pred$
is mono\-toni\-ca\-lly increasing.)
Given this, 
a number of problems were shown to be
represented as special cases of MSMP~\cite{msjb-cav13,msjm-aij17}.
Furthermore, it was shown that algorithms for finding minimal sets and
for enumerating minimal sets could be
devised~\cite{msjm-aij17,bendik-phd20} independently of specific
application domains.
Throughout this paper, when $\pred$ is unspecified, these algorithms
are \emph{uninstantiated}; 
and \emph{instantiated} otherwise, i.e., when an application
domain is known.

\paragraph{Measures of relative importance.}
A weighted voting game (WVG) is defined on a set $N$ of voters. With
each voter $i\in{N}$ one assigns a value $v_i\in\mbb{R}$. In addition,
a quota $q$ is given, with $q \leq \sum_{i\in{N}}v_i$. A coalition is any
subset of $N$. A winning coalition $S\subseteq{N}$ is such that
$\sum_{i\in{S}}v_i\ge{q}$. A coalition that is not a winning coalition
is a losing coalition. A minimal winning coalition is a winning
coalition such that any of its proper subsets is a losing coalition.

\begin{example}
  The notation
  $[7 ; 5, 5, 2, 1]$
  summarizes a WVG, with quota
  7, and four voters,
  each having respectively
  5, 5, 2, and 1 votes.
  The subset $\{1,3,4\}$ is an example of a winning coalition, whereas
  $\{1,3\}$ is a minimal winning coalition. Finally, $\{1,4\}$ is a
  losing coalition.
\end{example}

Since the 1940s~\cite{penrose-jrss46}, there has been interest in
assigning relative importance to voters of weighted voting games;
these measures are referred to as \emph{power indices}~\cite{machover-bk98}. 
In this paper, we 
focus on a few well-known power
indices, namely those of Shapley-Shubik~\cite{shapley-apsr54},
Banzhaf~\cite{banzhaf-rlr65} and Deegan-Packel~\cite{deegan-ijgt78}.

With each WVG, we associate a predicate $\winc:2^{N}\to\mbb{B}$, which
holds true for subsets of $N$ that represent winning coalitions.
Moreover, it is convenient to define a \emph{characteristic function}
(also referred to as a \emph{value function}), that maps subsets of
$N$ to the reals, as follows:
\begin{equation} \label{eq:pidx:cf}
  \cf(S) ~~ \coloneq ~~ \tn{ITE}(\winc(S),1,0)
\end{equation}
where $\tn{ITE}$ is the IF-THEN-ELSE operator.
(It should be underlined that the characteristic functions used in
applications other than weighted voting games often mimic the
characteristic function $\cf$ 
introduced in~\eqref{eq:pidx:cf}~\cite{hunter-aij10,amgoud-ijcai17,gusev-omega20,kimelfeld-sigmod23,lhms-corr24}.)

Given a voter $i\in{N}$ and a coalition $S\subseteq{N}$, the
difference in the value of the characteristic function due to voter
$i$ is given by,
\begin{equation} \label{eq:def:delta}
  \diff{i}(S) ~~ \coloneq ~~ \cf(S)-\cf(S\setminus\{i\})
\end{equation}

Existing measures of relative importance of a voter
$i\in{N}$ (e.g.,~\cite{shapley-apsr54,banzhaf-rlr65,deegan-ijgt78})
analyze all possible coalitions $S\subseteq{N}$.
For each coalition $S\subseteq{N}$, one accounts for the contribution
of $i$ for the coalition, i.e., $\diff{i}(S)$, weighted by a factor
$\coef(S)$, that depends on the power index being considered.
As a result, the general definition of a power index becomes:
\begin{equation} \label{eq:def:pidx}
\sv(i) ~~ \coloneq ~~ \sum\nolimits_{S\subseteq{N}}\coef(S)\times\diff{i}(S)
\end{equation}

The actual definitions of $\coef$ and $\diff{i}$ depend on the power
index considered, and will be revisited later in the paper.

\paragraph{Running examples.}
To illustrate the concepts introduced in the paper, we will consider
two running examples.

\begin{runexmpl}[Dominating sets]\label{runex:01}
  Given an undirected graph $G=(N,E)$, a dominating set is a subset
  $D \subseteq N$ 
  such that any vertex in $N$ is in $D$ or it is
  adjacent to a vertex in $D$.
  A minimal dominating set is a dominating set such that any of its
  proper subsets is not a dominating set. A
  minimum dominating set is a dominating set of the smallest size,
  and its size is known as the domination number of $G$, $\gamma(G)$.
  The decision problem of determining whether $\gamma(G)$ does not exceed
  a given value is a well-known NP-complete
  problem~\cite{garey-bk79}. 
  Moreover, dominating sets have been extensively studied in computer
  science, finding a wide range of practical
  applications~\cite{haynes-bk23}.
  We are interested in ranking the vertices of $G$ in terms of their
  relative importance for graph domination.

  In this context, the predicate $\dset:2^{N}\to\mbb{B}$ partitions the power set of $N$ into the subsets that
  are dominating sets and those that are not. 
  This predicate can be formally defined as
  $\dset(S) \coloneq \forall(i\in{N}).(i \in S \lor \exists (\{i, j\} \in E). j
  \in S)$, with $S \subseteq N$. 
  Clearly, $\dset$ is monotonically increasing, since any superset of
  a dominating set is also a dominating set. 

  \begin{figure}[t] 
   \centering
   \input{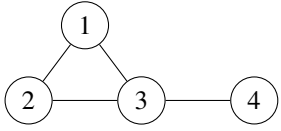}
   \caption{Graph for running example~\ref{runex:01}}\label{fig:domgraph}
  \end{figure}

  We will consider the graph shown in Figure ~\ref{fig:domgraph},
  where $N = \{1,2,3,4\}$.
  In this graph, $\{1,2,3\}$ is a dominating set,
  but $\{1,2\}$ is not, as it does not contain vertex $4$ nor any
  vertex adjacent to it. So, $\dset(\{1,2,3\}) = 1$ and $\dset(\{1,2\})=0$.
  In addition, there are 3 minimal dominating sets: 
  $\{1,4\}$,$\{2,4\}$ and $\{3\}$ (the only minimum one).
\end{runexmpl}

\begin{runexmpl}[Sardaukar training] \label{runex:02}
  In F.\ Herbert's Dune universe~\cite{dune-bk65}, the Sardaukar
  soldiers (SarS) are renowned for their fighting skills, but also for
  their incredibly harsh training. 
  The SarS training is composed of a number of extremely challenging
  physical exercises, each of which each SarS trainee is expected to
  master within a fixed time slot. When a SarS trainee fails to master
  exercise $i\in{N}$ in the allocated time slot, he/she is penalized
  with $p_i$ points. Moreover, when the sum of penalties of a SarS
  trainee exceeds a point threshold of $\Psi$, i.e.,\ the point
  threshold of no return, then the SarS trainee is automatically
  decommissioned in a rather hazardous fashion, i.e., he/she is
  terminated, thus justifying in part why about 50\% of SarS trainees
  fail to reach the age of 11. 
  (In this case, 
  $\sum_{i\in{N}}p_i\ge\Psi$.)
  We are interested in picking the first $K$
  exercises that are the most important for the decommissioning of
  SarS trainees.
  
  Given a set $S\subseteq{N}$ of exercises failed by a trainee, the
  predicate $\send:2^{N}\to\mbb{B}$ holds true if the sum of penalties
  of the failed exercises (in $S$)
  is no less than $\Psi$, i.e.,\ if the
  trainee is terminated. 
  $\send$ is monotonically
  increasing.
  
  We consider a concrete example where SarS trainees are subjected
  to six exercises, i.e., $N = \{1,...,6\}$, with respective penalties
  $\langle10,6,4,2,2,1\rangle$ and a threshold $\Psi=16$.
  For example, $\send(\{1,3,5\})=1$, given that $p_1=10,p_3=4,p_5=2$.
  The minimal sets of exercises that suffice for terminating a
  trainee are: $\{ \{1,2\}, \{1,3,4\}, \{1,3,5\} \}$. 
  Notice that exercise 6 is not included in any minimal set of
  exercises sufficient for a trainee to be decommissioned.
\end{runexmpl}

Observe that
the second running example
can be viewed as a disguised weighted voting
problem~\cite{machover-bk98}. Furthermore, the same example 
could also be related with systems of point-based penalties
associated with drivers' licenses, which are used in different
countries.

\section{Related Work}

\subsection{Measures of Importance}

\paragraph{Voting power.}
The measure of voter power in a weighted voting game 
is commonly conveyed through the use of \emph{power indices}, which
have been studied since the 1940s~\cite{penrose-jrss46}. The
connection of power indices with game theory was first studied in the
1950s~\cite{shapley-apsr54}.
Since then, different power indices have been
investigated~\cite{banzhaf-rlr65,johnston-ep78,deegan-ijgt78,holler-je83}.
Examples of more recent work on weighted voting games
include~\cite{andjiga-msh03,elkind-bk12,brandt-bk16,felsenthal-ho16,moretti-aor22}.
Generalizations of weighted voting games are also referred to as
characteristic function games~\cite{elkind-bk12}.

\paragraph{Inconsistent knowledge bases.}
Given an inconsistent knowledge base, one goal is to assign relative
importance, regarding the knowledge base inconsistency, to the formulas
in the knowledge base.
This line of research has been investigated since the mid
2000s~\cite{hunter-kr06,hunter-aij10,jabbour-ijcai23}.

\paragraph{Intensity of attacks in argumentation.}
Building on the application of measures of importance in inconsistent
knowledge bases, more recent work~\cite{amgoud-ijcai17} proposed their
use in argumentation, concretely in measuring the intensity of attacks.

\paragraph{Vertex cover.}
%
In the case of the well-known graph problem of (minimum or minimal)
vertex cover, recent work~\cite{gusev-omega20,gusev-ejor23} proposed the computation of 
measures of relative importance, in 
addition to relating vertex covers with different practical
applications.

\paragraph{Database management.}
In recent years, several applications of measures of importance have
been studied in the domain of database management. These have been
recently reviewed~\cite{kimelfeld-sigmod23}.

\paragraph{Explainable AI.}
For more than a decade, measures of relative importance have been
studied in the context of assigning influence to features given some
ML model prediction~\cite{kononenko-jmlr10,kononenko-kis14}. This line
of research has become highly visible with the proposal of
SHAP~\cite{lundberg-nips17}.
Unfortunately, the originally proposed measures of relative importance
exhibit several shortcomings~\cite{msh-cacm24,hms-ijar24}, which
motivated a stream of recent works on the
topic~\cite{ignatiev-corr23a,izza-aaai24,ignatiev-sat24,lhms-corr24,lhams-corr24}. 
These recent works build on ongoing research in logic-based
explainable AI~\cite{msi-aaai22,ms-rw22,darwiche-lics23}.

\subsection{Minimal Sets Over a Monotone Predicate (MSMP)}

Uninstantiated algorithms for computing minimal sets were first
discussed in the context of model
checking~\cite{bradley-fmcad07,bradley-fac08}. This initial work was
later extended to the domain of inconsistency
analysis~\cite{msjb-cav13,msm-ijcai20}, with the MSMP problem studied
in much greater detail soon
after~\cite{msjb-cav13,slaney-ecai14,jms-aij16,msjm-aij17,berryhill-sat18,bendik-phd20,berryhill-phd20,bendik-tacas21,rodler-kr21,bendik-lmcs22,mmms-icaps23}. 
To our best knowledge,
measures of importance have not been
considered in the context of MSMP.

\section{Uninstantiated Measures of Importance}

This section introduces Uninstantiated Measures of Importance
(\acronpl), by building on measures of importance used in
several domains:
a priori voting
power~\cite{shapley-apsr54,banzhaf-rlr65,deegan-ijgt78,johnston-ep78,holler-je83},
inconsistent knowledge bases~\cite{hunter-kr06,hunter-aij10}, and
argumentation frameworks~\cite{amgoud-ijcai17}, among others.

\paragraph{Minimal sets \& minimal hitting set duality.}
A Minimal Set for a (monotonically increasing) Predicate $\pred$ (MSP)
is any subset-minimal set $M\subseteq{N}$ such that $\pred(M)$ holds.
Predicate $\msp:2^{N}\to\mbb{B}$ holds on set $M\subseteq{N}$ if
$\pred(M)$ holds, and $M$ is subset-minimal, i.e.,
\[
\msp(M) \hspace{0.5em} \coloneq \hspace{0.5em} \pred(M)\land\forall(M'\subsetneq{M}).\neg\pred(M')
\]
A Minimal Break for a (monotonically increasing) Predicate $\pred$
(MBP) is any subset-minimal set $B\subseteq{N}$ such that
$\neg\pred(N\setminus{B})$ holds. Predicate $\mbp$ holds on set  
$B\subseteq{N}$ if $\neg\pred(N\setminus{B})$ holds, and $B$ is
subset-minimal, i.e.,
\[
\mbp(B) \hspace{0.5em} \coloneq \hspace{0.5em} \neg\pred(N\setminus{B})\land\forall(B'\subsetneq{B}).\pred(N\setminus{B'})
\]
Since $\pred$ is monotonically increasing, then the condition for
subset-minimality can be rewritten as follows:
\begin{align}
  \msp(M)& \hspace{0.5em} \coloneq \hspace{0.5em} \pred(M)\land\forall(t\in{M}).\neg\pred(M\setminus\{t\})
  \nonumber\\
  \mbp(B)& \hspace{0.5em} \coloneq \hspace{0.5em} \neg\pred(N\setminus{B})\land\forall(t\in{B}).\pred(N\setminus(B\setminus\{t\}))
  \nonumber
\end{align}
The modified definitions are vital for the practical
performance of algorithms for computing MSPs/MBPs. In contrast to the
original definitions, which require that all possible subsets be
analyzed, the modified definitions only require a number of subsets
linear on the size of the target set to be analyzed.
Moreover, MBPs are the complements of maximal sets for the
complemented predicate. A natural analogy are MUSes, MCSes and MSSes
in the case of inconsistent formulas~\cite{msm-ijcai20}. 

The set of MSPs is defined by $\mbb{M}=\{X\subseteq{N}\,|\,\msp(X)\}$.
Similarly, the set of MBPs is defined by
$\mbb{K}=\{X\subseteq{N}\,|\,\mbp(X)\}$.
These definitions can be restricted to the minimal sets that contain a
specific element $i\in{N}$, in which case a subscript $i$ is used,
i.e.,\ either $\mbb{M}_i$ or $\mbb{K}_i$.

Building on Reiter's seminal work~\cite{reiter-aij87}, the following
minimal hitting set (MHS) duality property is
well-known~\cite{slaney-ecai14}:

\begin{prop}
  $M\subseteq{N}$ is an MSP iff it is a minimal hitting set (MHS) of
  $\mbb{K}$.
  Also,
  $B\subseteq{N}$ is an MBP iff it is a minimal hitting set (MHS) of
  $\mbb{M}$.
\end{prop}

\paragraph{Computing MSPs/MBPs in practice.}
Given the definitions of MSP and MBP, 
the results and algorithms
devised for MSMP 
also hold in the case of
MSPs/MBPs.
This includes different algorithms for finding a minimal set,
including the well-known
Deletion~\cite{bakker-ijcai93},
Dichotomic~\cite{sais-ecai06},
QuickXplain~\cite{junker-aaai04} and
Progression~\cite{msjb-cav13} algorithms (among others), but also the
enumeration of minimal sets, including the well-known
MARCO algorithm~\cite{lpmms-cj16}, or later
improvements~\cite{bendik-phd20}.
This way, any problem that can be
formulated as the problem of computing one MSP/MBP or of enumerating
MSPs/MBPs can be solved using any existing MSMP algorithms.

\paragraph{Minimal sets as explanations.}
Each minimal set $S \subseteq N$ such that $\pred(S)$ holds is an
\emph{explanation} for $\pred$ to hold given $N$, in that $S$ is
sufficient for $\pred$ to hold, and $S$ is also (subset-) minimal.
In the context of (logic-based) XAI~\cite{msi-aaai22}, each abductive
explanation is a minimal set that is sufficient for a
monotonically increasing predicate, and that is also irreducible;
hence it is an explanation according to the previous definition.
However, as this section illustrates, the concept of explanation, as
used in XAI, also finds many other practical uses.

\paragraph{Measures of importance.}
Although the general MSMP framework proposed in earlier
works 
was analyzed in some detail, covering a number of important properties and
also algorithms for the computation and enumeration of (subset- or
cardinality-) minimal sets, what has not been studied are measures of
importance for the elements of $N$, regarding whether or not the
predicate $\pred$ holds.
(The existing exceptions for specific applications include 
those mentioned in the related work section,
but each one specific to its application
domain.)
We now show that several well-known measures of importance can be
defined in the 
general 
case of monotonically increasing predicates, and so are
applicable to a wide range of practical domains.

Given the characteristic function used in the case of power indices
(see~\eqref{eq:pidx:cf}), but also in other domains, the following
characteristic function is proposed:
\begin{equation}
  \cf(S) ~~ \coloneq ~~ \tn{ITE}(\pred(S),1,0)
\end{equation}
which can be stated, alternatively, as follows:%
\footnote{%
The second formulation was discussed in earlier
work~\cite{gusev-omega20,gusev-ejor23} in the context of vertex cover.
Also, the proposed characteristic function mimics exactly the ones used
in specific
domains~\cite{shapley-apsr54,hunter-aij10,amgoud-ijcai17,gusev-omega20,kimelfeld-sigmod23,lhms-corr24}.
}
$\cf(S)\coloneq\tn{ITE}(\exists(Z\in\mbb{M}).{Z}\subseteq{S},1,0)$.
Given that the codomain of $\cf$ is $\mbb{B}$ and that $\cf$ is
monotonically increasing (because $\pred$ is monotonically
increasing), then we are effectively reformulating MSMP as a
\emph{simple game}~\cite{elkind-bk12} with the goal of assigning
relative importance to the elements of $N$.

Using~\eqref{eq:def:delta}, we define
$\Delta(S)=\sum_{i\in{S}}\left(\Delta_i(S)\right)$, thus denoting the 
\emph{relative influence} of $S\subseteq{N}$.
Moreover, it is the case that $\Delta_i(S)\ge0$, since $\cf$ is
monotonically increasing.
Given that $\cf(S)\in\{0,1\}$, and that $\cf$ is monotonically increasing,
then one must either have $\Delta_i(S)=1$ or $\Delta_i(S)=0$. Finally,
for $\Delta_i(S)=1$, it must be the case that $\cf(S)=1$ and
$\cf(S\setminus\{i\})=0$.

An element 
$i \in N$ 
is \emph{critical} for a set $S\subseteq{N}$ to
be a set for $\pred$ if,%
\footnote{%
The concept of \emph{critical} element was already present in
Shapley\&Shubik's work~\cite{shapley-apsr54}. In other works it was
referred to as a \emph{swing}
element~\cite{shapley-mor79,lucas-prm83}, but also as a
\emph{decisive}
element~\cite{banzhaf-rlr65,lucas-prm83,machover-bk98,andjiga-msh03}, 
or as a \emph{marginal} element~\cite{lucas-prm83}.
These elements are, directly or indirectly, instrumental for the
definition of a panoply of power indices studied in the case of a
priori voting power~\cite{machover-bk98,andjiga-msh03}.}
\[
\critic_s(i,S) \hspace{0.5em} := \hspace{0.5em} \pred(S)\land\neg\pred(S\setminus\{i\})
\]

Thus, we have the following immediate result:
\begin{prop}
  Given the definition of $\cf$, $\Delta_i(S)=1$ if and only if
  $\critic_s(i,S)$.
\end{prop}
Essentially, $\Delta_i(S)=1$ for the elements 
$i \in N$
that are
critical for $S$ to be a set for $\pred$.
Furthermore, we could also consider an element 
$i \in N$ 
to be 
critical for a set $S\subseteq{N}$ to be a break for $\pred$.
Critical elements have also been indirectly considered in later
works~\cite{hunter-kr06,hunter-aij10,amgoud-ijcai17}, due to the
choice of characteristic function.
The following result follows from the definition of minimal sets and
critical element:
\begin{prop}
  For an MSP $M\subseteq{N}$, each element of $M$ is critical for $M$
  to be a set for $\pred$. 
  For an MBP $B\subseteq{N}$, each element of $B$ is critical for $B$
  to be a break for $\pred$. 
\end{prop}

Taking into consideration the definitions above, and the fact that the
proposed characteristic function mimics the one used in specific
domains, we can now redefine all of the best-known power indices in
the case of monotonically increasing predicates. These will be
referred to as Uninstantiated Measures of Importance (UMIs), to reflect
the framework in which the indices are defined.

An (uninstantiated) measure of importance $\im$ is a mapping from the
elements of $N$ to the reals, $\im:N\to\mbb{R}$. Different \acronpl
can be envisioned, and so for a given \acron $t$, $\im$ is qualified
with $t$, i.e.,\ $\imn{t}$.
Thus, the Shapley-Shubik~\cite{shapley-apsr54} ($\imn{S}$),
Banzhaf~\cite{banzhaf-rlr65} ($\imn{B}$), 
and
Deegan-Packel~\cite{deegan-ijgt78} ($\imn{D}$) 
\acronpl are defined as follows:
\begin{align}
  \imn{S}(i) & := 
  \sum\nolimits_{S\subseteq{N}\land\critic_s(i,S)}
  \left(\sfrac{1}{\left(|N|\times\binom{|N|-1}{|S|-1}\right)}\right)
  \nonumber\\[1.5pt]
  \imn{B}(i) & :=
  \sum\nolimits_{S\subseteq{N}\land\critic_s(i,S)}
  \left(\sfrac{1}{2^{|N|-1}}\right)
  \nonumber\\[1.5pt]
  \imn{D}(i) & :=
  \sum\nolimits_{S\in\mbb{M}_i}
  \left(\sfrac{1}{(|S|\times|\mbb{M}|)}\right)
  \nonumber
\end{align}
where the values of $\coef$ depend on each 
case, 
as introduced
in~\eqref{eq:def:pidx}.
Furthermore, $\imn{B}$ can optionally be normalized~\cite{shapley-mor79} 
so that its sum over all the elements of $N$ is 1 
(what holds for both $\imn{S}$ and $\imn{D}$). 
We will refer to this normalized version as $\imn{Bn}$.

As a brief explanation, for a given element $i \in N$, the Shapley-Shubik index can be defined as the fraction of the permutations of $N$ in which $i$ is \emph{pivotal}.
Given a permutation of the elements of $N$, the pivotal element is the first element that, together with the previous ones, makes the predicate to hold.
On the other hand, the Banzhaf UMI represents the fraction of times an element $i\in N$ is critical among all the subsets containing a critical element.
Alternatively, Deegan-Packel assigns a relative importance only focusing on the minimal sets containing the element $i$.

\subsection{Case Studies}

This section shows how measures of importance can be computed for the
two running examples of the paper.

\paragraph{Dominating sets.}

The UMIs described 
before 
can be instantiated to measure the importance
of each vertex in dominating the graph considered in the first running example. 

Recall that the set of vertices is $N = \{1,2,3,4\}$ and the minimal dominating sets are $\mbb{M} = \{ \{1,4\}, \{2,4\}, \{3\} \}$.
In this concrete case, the characteristic function is $\cf(S) \coloneq \tn{ITE}(\dset(S),1,0)$, with $S \subseteq N$.

The measures of importance are shown in Table~\ref{tab:runex:01} 
(rounded to three decimal places). 
\begin{table}[t]
  \begin{center}
    \renewcommand{\arraystretch}{1.25}
    \renewcommand{\tabcolsep}{0.5em} 
    \begin{tabular}{cccccc} \toprule
      \multirow{2}{*}{UMI} &
      \multicolumn{4}{c}{Vertex $\,i$}
      \\  \cline{2-5}
      & 1 & 2 & 3 & 4 
      \\
      \toprule
      $\imn{S}(i)$ & 0.083 & 0.083 & 0.583 & 0.250 
      \\
      $\imn{B}(i)$ & 0.125 & 0.125 & 0.625 & 0.375
      \\
      $\imn{Bn}(i)$ & 0.100 & 0.100 & 0.500 & 0.300
      \\
      $\imn{D}(i)$ & 0.167 & 0.167 & 0.333 & 0.333 
      \\
      \bottomrule
    \end{tabular}
  \end{center}
  \caption{UMIs for~\cref{runex:01} } \label{tab:runex:01}
\end{table}
As can be observed, 
Shapley-Shubik and Banzhaf yield similar results:
3 is deemed the most relevant vertex, followed by 4, and 1 and 2 tie as the least important ones. 
In contrast, Deegan-Packel assigns the same importance to vertices 3 and 4.

To illustrate the previous definitions, vertex 3 is critical for $D = \{1,2,3\}$ to be a dominating set, 
since $D$ is a dominating set but $D \setminus \{3\} = \{1,2\}$ is not.
However, vertex 2 is not critical for $D$ since $D \setminus \{2\} = \{1,3\}$ is still a dominating set.
So, $\critic_s(3,\{1,2,3\})$ holds and $\critic_s(2,\{1,2,3\})$ does not.

The computation of the Shapley-Shubik and Banzhaf indices depends on the sets for which a given vertex $i \in N$ is critical. 
As an example, vertex 4 is critical for three sets: $\{1,2,4\}$, $\{1,4\}$ and $\{2,4\}$.
Hence, $\{ S \subseteq N \land \critic_s(4,S) \} = \{\{1,2,4\}, \{1,4\}, \{2,4\}\}$.
In this case, the Shapley-Shubik value is computed as 
$\imn{S}(4) = (\sfrac{1}{(4\times\binom{3}{2})}) + (\sfrac{1}{(4\times\binom{3}{1})}) + (\sfrac{1}{(4\times\binom{3}{1})}) = \sfrac{3}{12} = 0.250$.
The Banzhaf index is $\imn{B}(4) = \sfrac{1}{2^3} + \sfrac{1}{2^3}  + \sfrac{1}{2^3}  = \sfrac{3}{8} =  0.375$.
This value is normalized as $\imn{Bn}(4) = 0.300$ to achieve a total sum of 1 accross all the elements of $N$.
On the other hand, Deegan-Packel only takes minimal sets into account. 
The minimal dominating sets containing vertex 4 are $\mbb{M}_4 = \{\{1,4\}, \{2,4\}\}$. 
Also, $| \mbb{M} |= 3$. So, $\imn{D}(4) = \sfrac{1}{(2 \times 3)} + \sfrac{1}{(2 \times 3)}   = \sfrac{1}{3} \approx 0.333$.

\paragraph{Sardaukar training.}
For~\cref{runex:02}, it is clear that failing exercise 6 is never
critical for a trainee to be terminated. Similarly, only if exercise 1
is failed can a trainee be terminated.
For example, it is plain to conclude that
$\critic_s(1,\{1,2,3,4,5,6\})$ holds. However,
$\critic_s(1,\{1,4,5,6\})$ does not hold. 
For each exercise $i$, we can find the sets $S\subseteq{N}$ for which
$i$ is critical. As a result, we can compute the UMIs proposed in the 
previous section. 

The results are summarized in~\cref{tab:runex:02}.
As noted earlier, for Deegan-Packel only the minimal sets are
considered; in this case each exercise in each minimal set is also
critical for that set.
The values for exercise 6 should be unsurprising. As argued
earlier, exercise 6 is referred to as \emph{irrelevant} in
XAI~\cite{msh-cacm24}, or as \emph{dummy} in a priori voting 
power~\cite{lucas-prm83}.
\begin{table}[t]
  \begin{center}
    \renewcommand{\arraystretch}{1.25}
    \renewcommand{\tabcolsep}{0.5em} 
    \begin{tabular}{ccccccc} \toprule
      \multirow{2}{*}{UMI} &
      \multicolumn{6}{c}{Exercise \#$\,i$}
      \\  \cline{2-7}
      & 1 & 2 & 3 & 4 & 5 & 6
      \\
      \toprule
      $\imn{S}(i)$ & 0.617 & 0.200 & 0.117 & 0.033 & 0.033 & 0.000
      \\
      $\imn{B}(i)$ & 1.000 & 0.455 & 0.273 & 0.091 & 0.091 & 0.000
      \\
      $\imn{Bn}(i)$ & 0.524 & 0.238 & 0.143 & 0.048 & 0.048 & 0.000 
      \\
      $\imn{D}(i)$ & 0.389 & 0.167 & 0.222 & 0.111 & 0.111 & 0.000
      \\
      \bottomrule
    \end{tabular}
  \end{center}
  \caption{UMIs for~\cref{runex:02}} \label{tab:runex:02}
\end{table}

Finally, 
as already observed in the first running example, 
the relative importance of the
different exercises is not always the same. For this 
example,
and for Shapley-Shubik and Banzhaf, the obtained relative importances
are the same. However, for Deegan-Packel 
it changes,
with exercise 3 deemed more important than exercise
2. It is debatable which ranking of exercises should be deemed the
most adequate. However, in different domains of application,
the Shapley-Shubik and Banzhaf indices find a much larger range of
uses.

\subsection{Example Application Areas}

Besides the two case studies and the application domains already
discussed in the paper, both MSMP and measures of importance for MSMP
find a wide range of applications.
\cite{msjm-aij17,msm-ijcai20} study several examples related with
logic formulas. One example are minimal unsatisfiable subsets and
minimal correction subsets, but one can also account for their many 
generalizations, including fragments of first order logic.
Moreover, \cite{eiter-jelia02,eiter-dam08,vera-licona-sjdm17}
discuss other related examples. In addition, \cite{gusev-omega20,gusev-ejor23}
lists several practical uses related with set covering. The measures
of importance proposed in this paper can be applied to \emph{any} of
these examples.
Furthermore, examples from other different domains can also be
identified. This section briefly discusses some of these additional
examples. 

Arguably, UMIs can be devised for the following computational problems:
\footnote{To the best of our knowledge, measures of importance have
not been studied for any of these computational problems.}
\begin{enumerate}
\item Model-based diagnosis~\cite{reiter-aij87};
\item Inconsistent linear
  inequalities~\cite{vanloon-ejor81,chinneck-jc91};
\item Axiom pinpointing in description
  logics~\cite{baader-jlc10,amms-sat15,kazakov-ijcar18}; 
\item Consistent query
  answering~\cite{kolaitis-sat19,kolaitis-icde22}; 
\item Prime implicants (resp.\ implicates) given a term
  (resp.\ clause)~\cite{rymon-amai94,pimms-ijcai15};
\item Multigenome alignment~\cite{karp-soda11,karp-or13};
\item Metabolic networks~\cite{klamt-bioinf12,klamt-bmcbioinf20};
\item Inconsistencies in biological
  networks~\cite{schaub-iclp08,schaub-tlp11}; 
\item Model reconciliation~\cite{previti-aaai21};
\item Generating sets in finite algebras~\cite{janota-jsc23}.
\end{enumerate}

The key observation is that, for all the applications listed above,
but also for many related applications, one targets the computation of
a minimal set subject to a monotonically increasing predicate. As a
result, the computation of relative measures of importance proposed in
this paper is also applicable to those applications.

\subsection{Exact Computation \& Approximation}

Given the reduction of measures of importance to simple games,
general complexity results apply~\cite{elkind-bk12}.
Furthermore, specific complexity results have also been studied~\cite{vandenbroeck-jair22,barcelo-jmlr23}.
In addition, approximate solutions have been studied in different
settings~\cite{jennings-aij08,tejada-cor09,jennings-mgs12,sais-cor21,ignatiev-sat24}.

\section{Conclusions \& Research Directions} \label{sec:conc}

Monotone predicates are ubiquitous in different domains of
computing~\cite{msjb-cav13,slaney-ecai14,msjm-aij17}, but also in
other fields~\cite{shapley-apsr54,machover-bk98,gusev-omega20,moretti-aor22}.
Over the years, researchers have studied measures of relative
importance of elements in different contexts, that include, among
others, a priori voting power, inconsistency of knowledge bases, 
database management and explainability. 
This paper shows that such measures 
can be
computed in a much wider range of domains than previously understood.
Furthermore, the paper argues that approaches for computing such
measures of relative importance 
depend not on the exact problem being solved, but instead on the
properties of such problem.

Motivated by the results in this paper, several research directions
can be envisioned, and many more should be expected. For example, for
some domains of application, computing measures of importance
represents a novel area of research. One example is the computation of
prime implicants given a specific term. Similarly, computing relative 
measures of importance for inconsistent linear programs represents
another novel area of research. Additional domains of application are
discussed in the paper. Furthermore, the approximate computation of
relative measures of importance, which has been studied in recent
work can also be applied to other domains, including those discussed in this
paper.

\section{Acknowledgements}
This work is partially supported by the Spanish Government under
grants PID2023-152814OB-I00, PID2022-141746OB-I00 
and TED2021-131938B-I00, and by ICREA
starting funds.

\newtoggle{mkbbl}

\settoggle{mkbbl}{false}

\iftoggle{mkbbl}{
  \bibliography{team,refs,pidx,qna,xai,xtra,other}
}{

}
\let\addcontentsline\oldaddcontentsline

\end{document}